\documentclass[letterpaper, preprint, paper,11pt]{AAS}	% for preprint proceedings

\usepackage{bm}
\usepackage{amsmath}
\usepackage[colorlinks=true, pdfstartview=FitV, linkcolor=black, citecolor= black, urlcolor= black]{hyperref}
\usepackage{overcite}
\usepackage{footnpag}			      	% make footnote symbols restart on each page

\usepackage{amssymb,amsfonts}
\usepackage{multirow}

\usepackage{subcaption}

\usepackage{xcolor}

\usepackage{soul}

\PaperNumber{26-723}

\begin{document}

\title{Multi-Axis Selective Decoupling Framework for Free-Floating Space Manipulators}

\author{Daegyun Choi\thanks{Postdoctoral Researcher, Department of Aerospace Engineering \& Engineering Mechanics, University of Cincinnati, Cincinnati, OH 45221, USA.} %,  
\ and Donghoon Kim\thanks{Associate Professor, Department of Aerospace Engineering \& Engineering Mechanics, University of Cincinnati, Cincinnati, OH 45221, USA.}
% ,
% \ and J.Q. Public\thanks{Title, department, affiliation, postal address.}
}

\maketitle{}

\begin{abstract}
Free-floating space manipulator systems face operational challenges due to dynamic coupling. While traditional reactionless manipulation eliminates base disturbances, enforcing the null-space projections across all rotational axes imposes excessive constraints on the system and drastically reduces the available workspace. To address this limitation, this work proposes a multi-axis selective decoupling framework that nullifies momentum transfer exclusively along mission-critical directions. By leveraging a directionally constrained sub-coupling matrix, the framework relaxes the null-space constraints. Numerical simulations confirm that this approach mathematically eliminates constrained directional disturbances along the targeted axes while preserving the remaining degrees of freedom, enabling the manipulator to execute task-space operations successfully.
\end{abstract}

\section{Introduction}
The rapid advancement of space technologies and increasing attention to space exploration have influenced a significant shift toward in-space servicing, assembly, and manufacturing. Space manipulator systems (SMSs) are key to this evolution, serving as the primary systems for complex missions, such as the repair of aging or faulty components, autonomous refueling, structural assembly, and active debris removal \cite{flores2014review}. As deployed orbital assets age and newly launched systems become increasingly sophisticated and expensive, the ability to extend their operational lifespans through robotic systems has gained paramount importance.

In SMSs, two distinct modes are typically considered: free-flying and free-floating \cite{papadopoulos2021robotic}. In the free-flying mode, the spacecraft base utilizes active attitude and position control, such as thrusters, reaction wheels, or control moment gyroscopes, to maintain a fixed or commanded base state during manipulator maneuvers. Conversely, the free-floating mode allows the base to move freely in response to the robotic arm's motion, governed by the conservation of momentum \cite{vafa1990dynamics}. While the free-flying mode offers a stable platform for precision tasks, the free-floating mode is often preferred for long-duration missions due to its fuel efficiency and the elimination of thruster plume contamination, which can degrade sensitive sensors or optical surfaces on the target satellite \cite{oda2000experiences}.

However, the free-floating mode introduces a significant challenge: the dynamic coupling between the robotic arm and the base spacecraft. Any movement of the robotic arm induces a reactive motion on the base, potentially perturbing its 6-degree-of-freedom (DOF) motion and hindering the mission objectives of the end-effector (EE) \cite{das2025understanding}. Historically, this base motion has been regarded as an undesired disturbance that must be minimized to preserve mission integrity. Excessive base drift can lead to the saturation of momentum exchange devices or may require frequent reaction-thruster-driven desaturation maneuvers, reducing mission life. 

To mitigate this, researchers have employed optimization techniques to minimize the overall magnitude of the base disturbance while maximizing the arm's manipulability \cite{misra2017task, lu2020trajectory}. However, by relying on scalar cost functions, such as the total magnitude of the base disturbance, these approaches overlook the directional characteristics of the base's reactive motion. This can introduce critical issues because even a small residual disturbance aligned with highly sensitive axes can trigger resonant vibrations in flexible appendages or hinder precise instrument pointing, compromising mission safety and success. To eliminate such risks, some studies have focused on reactionless manipulation, attempting to suppress the base disturbance across all body rotational axes by leveraging null-space projection \cite{yoshida2003engineering,james2016reactionless}. While this approach is appealing, enforcing a strict 3-DOF null-space projection may severely reduce the operational workspace \cite{nenchev1992analysis}, drastically limit the achievable EE motion \cite{dubowsky1993kinematics}, and frequently drive the manipulator into singularities \cite{papadopoulos1993dynamic}. This renders the system overly conservative and inflexible for complex servicing missions.

In practice, strictly preventing base motion across all three axes simultaneously is often unnecessary and overly restrictive. Due to asymmetric inertial properties, spacecraft and SMSs are exceptionally vulnerable to disturbances along specific directions, such as out-of-plane bending in large, flexible solar arrays \cite{na2014vibration} or line-of-sight (LOS) deviations in high-gain antennas \cite{nguyen2013adaptive}. In such cases, isolating the disturbance along one or two critical axes is sufficient, allowing the remaining axes to safely absorb the reactive momentum. Despite the practical need for selective axis decoupling, existing literature has mainly evaluated dynamic coupling through scalar metrics, such as the total magnitude of the disturbance \cite{xu1993measure, zhou2019dynamic, chhabra2025reconfigurable}, manipulability ellipsoid indices \cite{zhou2021dynamic}, and disturbance maps \cite{vafa1990dynamics}. While these metrics provide valuable geometric intuition for path planning by identifying areas of low- or high-intensity coupling, they are primarily utilized to evaluate the overall magnitude rather than the specific directionality of the base reaction. Consequently, there remains a critical research gap in developing directional insights and control strategies that isolate specific base reaction directions from the manipulator motion.

To bridge this gap, this study extends beyond magnitude-based coupling analysis to introduce a generalized framework for multi-axis selective decoupling in free-floating SMSs. By leveraging the properties of the inertia-weighted rotational coupling matrix, this study provides a new perspective on how individual joints contribute to directional base excitation. Rather than imposing a restrictive total-reactionless constraint, the proposed strategy systematically targets and nullifies momentum transfer exclusively along a user-defined task-critical subspace. By identifying and isolating joint configurations that eliminate disturbances solely along these sensitive directions, the framework relaxes the null-space constraints. This allows the manipulator to retain expanded operational workspaces, preserve redundant DOF for primary manipulation tasks, and actively avoid kinematic and dynamic singularities.

\section{Directional Dynamic Coupling and Selective Decoupling Framework}
This section introduces a comprehensive framework for analyzing and managing dynamic coupling in free-floating SMSs. By deriving the rotational coupling matrix and evaluating its directional characteristics, a multi-axis decoupling strategy is established to selectively mitigate base disturbances during manipulator operation.

\subsection{Conservation of Momentum and Rotational Coupling}
In the absence of external forces and torques, the linear and angular momenta of a free-floating SMS are conserved. While the arm's motion induces both translational and rotational motion of the base, the translational drift is a holonomic function of the joint configuration; thus, the base position is uniquely determined by the current joint angles \cite{nakamura1990nonholonomic}. In contrast, rotational coupling is non-holonomic, meaning the base attitude depends on the history of joint motions. This history-dependent behavior makes attitude stability a critical and challenging factor for mission success \cite{umetani1989resolved}. Consequently, this work focuses specifically on the rotational coupling effect. 

The angular momentum ${}^{N}{\bf l}$ about the system center of mass is given by \cite{yoshida2003engineering}:
\begin{equation}
    {}^{N}{\bf l} = {}^{N}H_{\omega\omega} {}^{N}\bm{\omega}_{B/N} +  {}^{N}H_{\omega q}\dot{{\bf q}},\label{eq:ang_mom}
\end{equation}
where ${}^{N}H_{\omega\omega}\in\mathbb{R}^{3 \times 3}$ is the system inertia matrix, ${}^{N}H_{\omega q}\in\mathbb{R}^{3 \times n}$ is the coupling matrix, ${}^{N}{\bm\omega}_{B/N}\in\mathbb{R}^{3}$ is the angular velocity of the base with respect to the inertial frame, and $\dot{\bf q}\in\mathbb{R}^n$ is the joint rate vector. The left superscript $N$ indicates that the components are expressed in the inertial frame. Note that the joint rate of each link is evaluated with respect to its preceding rigid body.

Assuming zero initial momentum, Eq.~\eqref{eq:ang_mom} yields the kinodynamic constraint mapping the joint-space velocity to the base rotational velocity:
\begin{equation}
    {}^{N}\bm{\omega}_{B/N} = - {}^{N}H_{\omega\omega}^{-1}  {}^{N}H_{\omega q}\dot{{\bf q}} = {}^{N}J_{rc}\dot{{\bf q}},
\end{equation}
where ${}^{N}J_{rc}\in\mathbb{R}^{3 \times n}$ represents the inertia-weighted joint-to-base rotational coupling Jacobian. To isolate the intrinsic dynamic coupling characteristics from the time-varying inertial orientation of the spacecraft, this work maps the operator into the base-fixed frame:
\begin{equation}
    J_{rc} = - H_{\omega\omega}^{-1}  H_{\omega q},
\end{equation}
where $H_{\omega\omega} = C_{BN} \ {}^{N}H_{\omega\omega}C_{BN}^T$ and $H_{\omega q} = C_{BN} \ {}^{N}H_{\omega q}$, with $C_{BN}$ denoting the direction cosine matrix from the inertial to the base frame.

\subsection{Multi-Axis Selective Decoupling Framework}
Existing studies have predominantly concentrated on eliminating base attitude disturbances by restricting the commanded joint rate to the null space of the full rotational coupling Jacobian $J_{rc}$. Although this approach effectively stabilizes the base during operation, enforcing zero disturbance across all rotational axes simultaneously imposes an over-constrained condition that unnecessarily restricts the manipulator's operational workspace. To mitigate this limitation, a multi-axis selective decoupling framework is proposed to extract and isolate momentum transfer exclusively along mission-critical directions, preserving remaining DOF for the primary manipulation task.

For a free-floating SMS initially at rest, the base angular velocity $\bm\omega_{B/N}$ is governed by the conservation of angular momentum. This relationship maps joint rates to base angular velocity via the inertia-weighted rotational coupling Jacobian matrix:
\begin{equation}\label{eq:angvel}
    \bm\omega_{B/N} = J_{rc} \dot{\bf q}.
\end{equation}

To prevent momentum transfer to the base along targeted configurations, the rotational disturbances must be strictly nullified along specific directions requiring protection. Let the targeted protection space be spanned by a generalized directional constraint matrix $U_{c}\in\mathbb{R}^{3\times k}$:
\begin{equation}
    U_c = [{\bf u}_{c1} \ \vdots \ \cdots  \ \vdots \ {\bf u}_{ck}],
\end{equation}
where $k$ ($1 \le k \le 3$) represents the number of rotational axes to be decoupled, and ${\bf u}_{ci}\in\mathbb{R}^3$ are orthogonal unit vectors defining the vulnerable directions where rotational disturbances must be eliminated.

The necessary and sufficient condition to ensure zero base excitation exclusively along the defined directions is that the orthogonal projection of the base angular velocity onto the constraint space $\mathcal{S}(U_c)$ must strictly vanish. By projecting $\bm{\omega}_{B/N}$ from Eq.~\eqref{eq:angvel} onto $U_c$, the constrained disturbance angular velocity $\bm\omega_\text{sub}\in\mathbb{R}^k$ is derived as:
\begin{equation}
    \bm\omega_\text{sub} = U_c^T \bm\omega_{B/N} = U_c^T J_{rc} \dot{\bf q} = {\bf 0}.
\end{equation}
From this relationship, the directionally constrained sub-coupling matrix $J_\text{sub}\in\mathbb{R}^{k\times n}$ is defined as:
\begin{equation}
    J_\text{sub} \triangleq U_c^T J_{rc}.
\end{equation}
Consequently, the strict kinematic constraint derived from momentum conservation yields:
\begin{equation}\label{eq:const}
    J_{\text{sub}} \dot{\bf q} = {\bf 0}.
\end{equation}

To practically implement this selective decoupling strategy while simultaneously executing the primary task of the EE, a priority-based multi-task motion-planning optimization problem is formulated. Because base disturbance elimination along the targeted directions is critically sensitive, the selective decoupling condition in Eq.~\eqref{eq:const} is treated as a hard primary kinematic constraint. The primary regulation task of the EE is then handled via the remaining operational redundancy. 

The task-space reference velocity command ${\bf v}_{\text{cmd}} \in \mathbb{R}^3$ required to ensure asymptotic convergence of the regulation error is defined via a proportional control law \cite{marti2020multi}:
\begin{equation}
    {\bf v}_\text{cmd} = \dot{\bf x}_\text{des} + k_p({\bf x}_\text{des} - {\bf x}_{ee}),
\end{equation}
where $\dot{\bf x}_\text{des}$, ${\bf x}_\text{des}$, and ${\bf x}_{ee}$ denote the desired EE velocity, desired EE position, and current EE position, respectively, and $k_p>0$ is a scalar tracking gain. To map this demand into joint space, the commanded velocity must satisfy the differential kinematics relation:
\begin{equation}\label{eq:v_cmd}
    J_{g,t} \dot{\bf q}_{\text{cmd}} = {\bf v}_{\text{cmd}},
\end{equation}
where $J_{g,t}\in\mathbb{R}^{3 \times n}$ represents the translational generalized Jacobian matrix. 

To resolve the joint velocity distribution under strict adherence to the primary decoupling constraint in Eq.~\eqref{eq:const}, an optimization problem is structured to minimize the deviation from the primary task demands while minimizing the kinetic norm of the joint velocities. Recognizing that any valid joint command must reside entirely within the null space of the decoupling matrix, the joint rate vector is structurally constrained to the form $\dot{\bf q} = N_\text{sub} \dot{\bf q}_\text{aux}$, where $N_{\text{sub}} = I_{n \times n} - J_{\text{sub}}^\dagger J_{\text{sub}}$ represents the unique orthogonal null-space projection matrix of $J_\text{sub}$, $I_{n \times n}$ is the identity matrix, and $(\cdot)^\dagger$ denotes the Moore-Penrose pseudoinverse operator. 

The corresponding task-space tracking objective function is formulated in terms of the auxiliary joint rates as:
\begin{equation}\label{eq:opt_prob}
    \min_{\dot{\bf q}_{\text{aux}}} \mathcal{L}%(\dot{\bf q}_{\text{aux}}) 
    = \frac{1}{2} (J_{g,t} N_{\text{sub}} \dot{\bf q}_{\text{aux}} - {\bf v}_{\text{cmd}})^T (J_{g,t} N_{\text{sub}} \dot{\bf q}_{\text{aux}} - {\bf v}_{\text{cmd}}) + \frac{1}{2} \rho \dot{\bf q}_{\text{aux}}^T \dot{\bf q}_{\text{aux}},
\end{equation}
where $\rho > 0$ is a regularizing scalar to maintain minimum-norm regular solutions. Adopting a Lagrange multiplier vector $\bm{\lambda} \in \mathbb{R}^3$ for the exact tracking limit where $\rho \to 0$, the corresponding Lagrangian function for the constrained minimization is given by:
\begin{equation}\label{eq:lagrangian}
    \mathcal{H}%(\dot{\bf q}_{\text{aux}}, \bm{\lambda}) 
    = \frac{1}{2}\dot{\bf q}_{\text{aux}}^T \dot{\bf q}_{\text{aux}} + \bm{\lambda}^T (J_{g,t} N_{\text{sub}} \dot{\bf q}_{\text{aux}} - {\bf v}_{\text{cmd}}).
\end{equation}
The first-order necessary condition for optimality, evaluated by taking the partial derivative with respect to the auxiliary joint rate vector $\partial \mathcal{H} / \partial \dot{\bf q}_{\text{aux}} = {\bf 0}$, yields:
\begin{equation}\label{eq:first_order}
    \dot{\bf q}_{\text{aux}} = - (J_{g,t} N_{\text{sub}})^T \bm{\lambda}.
\end{equation}
Substituting the optimality condition in Eq.~\eqref{eq:first_order} back into the mapped differential kinematic task constraint yields:
\begin{equation}\label{eq:lambda_sub}
    - J_{g,t} N_{\text{sub}} (J_{g,t} N_{\text{sub}})^T \bm{\lambda} = {\bf v}_{\text{cmd}}.
\end{equation}
Assuming that the combined task-space matrix $J_{g,t} N_{\text{sub}}$ maintains full row rank, which physically implies that the secondary EE task does not conflict with the primary axis-decoupling constraint, solving Eq.~\eqref{eq:lambda_sub} explicitly for the Lagrange multiplier vector produces:
\begin{equation}\label{eq:lambda_sol}
    \bm{\lambda} = - [ (J_{g,t} N_{\text{sub}})(J_{g,t} N_{\text{sub}})^T ]^{-1} {\bf v}_{\text{cmd}}.
\end{equation}
Substituting the analytic solution of the multiplier vector from Eq.~\eqref{eq:lambda_sol} back into the first-order necessary condition in Eq.~\eqref{eq:first_order} produces the optimal minimum-norm auxiliary joint command:
\begin{equation}\label{eq:aux_final}
    \dot{\bf q}_{\text{aux}} = (J_{g,t} N_{\text{sub}})^T \left[ (J_{g,t} N_{\text{sub}})(J_{g,t} N_{\text{sub}})^T \right]^{-1} {\bf v}_{\text{cmd}}%= (J_{g,t} N_{\text{sub}})^\dagger {\bf v}_{\text{cmd}}
    .
\end{equation}

The final command law for the actual joints is obtained by projecting the optimal auxiliary vector $\dot{\bf q}_{\text{aux}}$ through the selective null-space matrix. Utilizing the structural definition $\dot{\bf q}_\text{cmd} = N_\text{sub} \dot{\bf q}_\text{aux}$ alongside the algebraic identity for projection operations where $N_{\text{sub}}(J_{g,t}N_{\text{sub}})^\dagger = (J_{g,t}N_{\text{sub}})^\dagger$, the final selective decoupling resolution planning law is derived as:
\begin{equation}
    \dot{\bf q}_\text{cmd} = (J_{g,t}N_\text{sub})^\dagger {\bf v}_\text{cmd}.
\end{equation}
The second-order sufficient condition is satisfied since the Hessian matrix of the Lagrangian function is the identity matrix, $\partial^2 \mathcal{H} / (\partial \dot{\bf q}_{\text{aux}} \partial \dot{\bf q}_{\text{aux}}^T) = I_{n \times n} \succ 0$. 

This analytical closed-form law guarantees mathematically exact tracking of the primary task command while simultaneously preserving zero momentum transfer to the spacecraft base exclusively along the mission-critical axes defined by $U_c$. The primary benefit of this framework is its flexibility to adapt dynamically to varying base disturbance reduction profiles while minimizing the unnecessary loss of functional manipulator's DOF. Depending on the dimensionality of the targeted base constraint, the framework systematically unifies and expands separate decoupling methodologies:
\begin{itemize}
    \item \textbf{Single-axis decoupling ($k=1$):} Protects a single critical axis of the base, such as isolating out-of-plane bending resonance for large, flexible solar arrays. The manipulator retains $n-1$ operational DOF to execute the task-space mission.
    \item \textbf{Two-axes decoupling ($k=2$):} Protects a critical planar subspace of the base. For instance, maintaining LOS stabilization for a communication antenna requires nullifying disturbances along both the pitch and yaw axes. In this scenario, $U_c$ is constructed with two directional orthogonal unit vectors, leveraging a 2-DOF 
    null-space projection and freeing $n-2$ operational DOF for task execution.
\end{itemize}

\section{Simulation Study}
\begin{figure}[!b]
    \centering
    \includegraphics[width=0.9\linewidth]{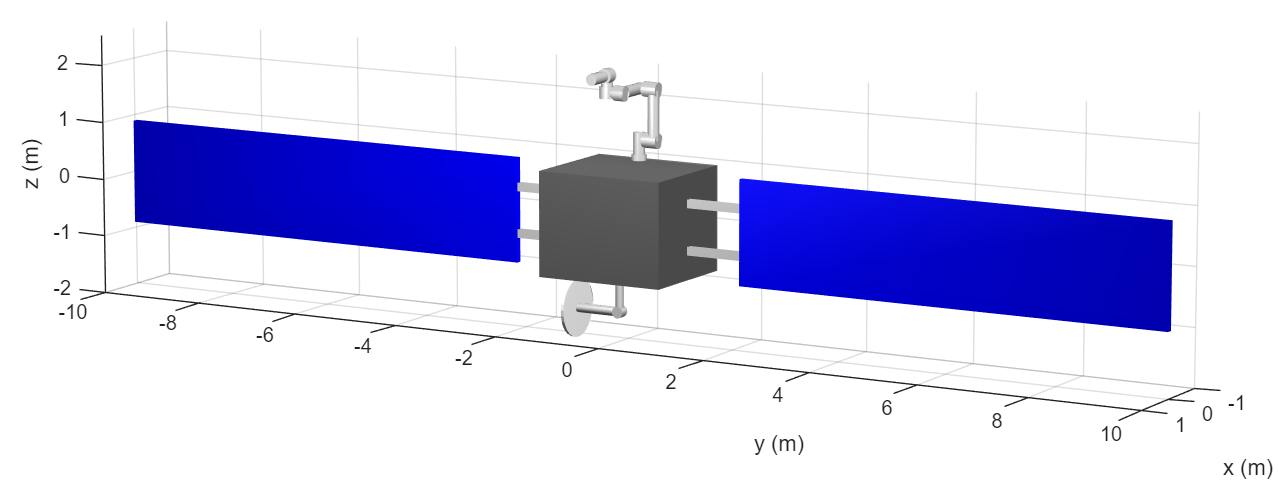}
    \caption{Visual Representation of ETS-VII}
    \label{fig:sms_config}
\end{figure}

To evaluate the performance and verify the mathematical validity of the proposed multi-axis selective decoupling framework, numerical simulation studies are conducted using a free-floating SMS. The Engineering Test Satellite VII (ETS-VII) serves as the physical reference model. The kinematic and dynamic properties of the base spacecraft and the 6-DOF robotic manipulator are structured based on the actual ETS-VII specifications \cite{yoshida2003engineering} and listed in Table~\ref{tab:sms_param}. The manipulator is mounted at $[-0.79, -0.29, 1]^T$~m from the center of the base, which is defined as the system origin $[0, 0, 0]^T$. Furthermore, critical structural components, such as solar arrays and a high-gain antenna, are included in the model (as shown in Figure~\ref{fig:sms_config}) to highlight realistic operational scenarios and the susceptibility of these appendages to reaction-induced base disturbances. 

\begin{table}[!b]
    \centering
    \caption{Kinematic and Dynamic Parameters of the Reference SMS Model}
    \label{tab:sms_param}
    \begin{tabular}{c|c|c|c}
        \hline
        Body    & Mass (kg) & Length (m)    & Inertia (kg$\cdot$m$^2$) \\ 
        \hline
        Base    & $2550.0$  & {--}          & diag([$6200$, $3540$, $7090$]) \\
        Link 1  & $35.0$    & $0.35$        & \multirow{6}{11em}{Calculated as solid cylinders with the given link length and a radius of 0.1~m \cite{Kanzawa2001}
        } \\
        Link 2  & $22.5$    & $0.87$        &  \\
        Link 3  & $21.9$    & $0.63$        &  \\
        Link 4  & $16.5$    & $0.26$        &  \\
        Link 5  & $26.0$    & $0.28$        &  \\
        Link 6  & $18.5$    & $0.53$        &  \\ 
        \hline
    \end{tabular}
\end{table}

To demonstrate the proposed framework's ability to protect specific mission-critical axes, three representative scenarios are considered:
\begin{itemize}
    \item \textbf{Case 1) Specific principal body-axis decoupling for protection from solar array bending mode ($k=1$)}: Large and flexible appendages, such as solar arrays, are vulnerable to bending mode excitation. This bending moment is transferred back to the base, disturbing the overall base attitude. Because the bending mode is often dominant along a single direction, the yaw axis of the base is selected as the protection axis (i.e., $U_c = [0, 0, 1]^T$), as shown in Figure~\ref{fig:sms_config}.
    \item \textbf{Case 2) Generalized directional decoupling ($k=1$)}: In practical scenarios, the vulnerable direction demanding protection may not align perfectly with the base's principal axes. This case demonstrates decoupling along a tilted direction relative to the yaw axis (i.e., $U_c=[0.5, 0, 0.866]^T$), emulating a solar array rotated to track the Sun.
    \item \textbf{Case 3) Multi-axis decoupling for antenna LOS maintenance ($k=2$)}: Maintaining a high-gain antenna's LOS toward a ground station requires suppressing base attitude disturbances that cause pointing errors and signal degradation. Assuming that the LOS must be stabilized while permitting rotations about the pitch axis, the roll and yaw axes are targeted for disturbance suppression (i.e., $U_c = [[1, 0, 0]^T\ \vdots \ [0, 0, 1]^T]$).
\end{itemize}

The simulation parameters used in this work are summarized in Table~\ref{tab:sim_param}.

\begin{table}[!b]
    \centering
    \caption{{Simulation Parameters}}
    \label{tab:sim_param}
    \begin{tabular}{c | c | c}
    \hline
        \multicolumn{2}{c|}{Parameter (unit)}                               & Value \\ 
        \hline
        \multirow[t]{6}{*}{All cases}   & Simulation time (s)               & $150$ \\
                                        & Time interval (s)                 & $0.1$ \\ 
                                        & Initial base states               & zeros \\
                                        & Initial joint rates               & zeros \\
                                        & Tracking gain                     & $0.05$ \\ 
                                        & Max. joint rate (deg/s)           & $5$ \\
        \multirow[t]{2}{*}{Case 1}      & Initial joint configuration (deg) & $[150,-20,-10,0,0,0]^T$ \\
                                        & Desired EE position (m)           & $[-1.8,-0.3,1.2]^T$ \\
        \multirow[t]{2}{*}{Case 2}      & Initial joint configuration (deg) & $[90,-30,0,0,0,0]^T$ \\
                                        & Desired EE position (m)           & $[-1.5,-0.3,1.2]^T$ \\
        \multirow[t]{2}{*}{Case 3}      & Initial joint configuration (deg) & $[180,0,-20,0,0,0]^T$ \\
                                        & Desired EE position (m)           & $[-1.5,0.1,1.1]^T$ \\
        \hline
    \end{tabular}
\end{table}

\begin{figure}[!t]
    \centering
    \begin{subfigure}[b]{0.49\textwidth}
        \centering
        \includegraphics[width=\textwidth]{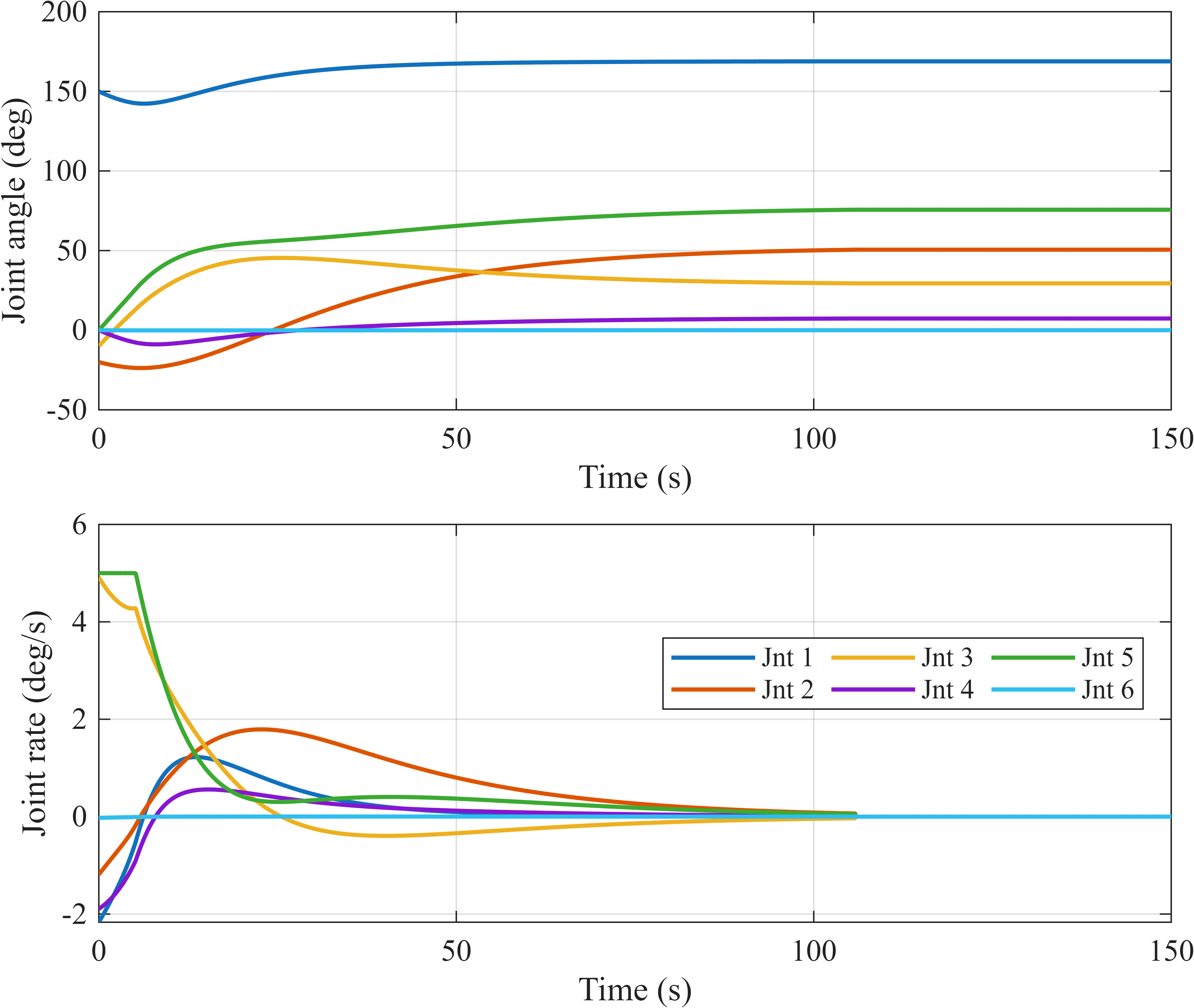}
        \label{fig:case1_jnt}
    \end{subfigure}
    \begin{subfigure}[b]{0.49\textwidth}
        \centering
        \includegraphics[width=\textwidth]{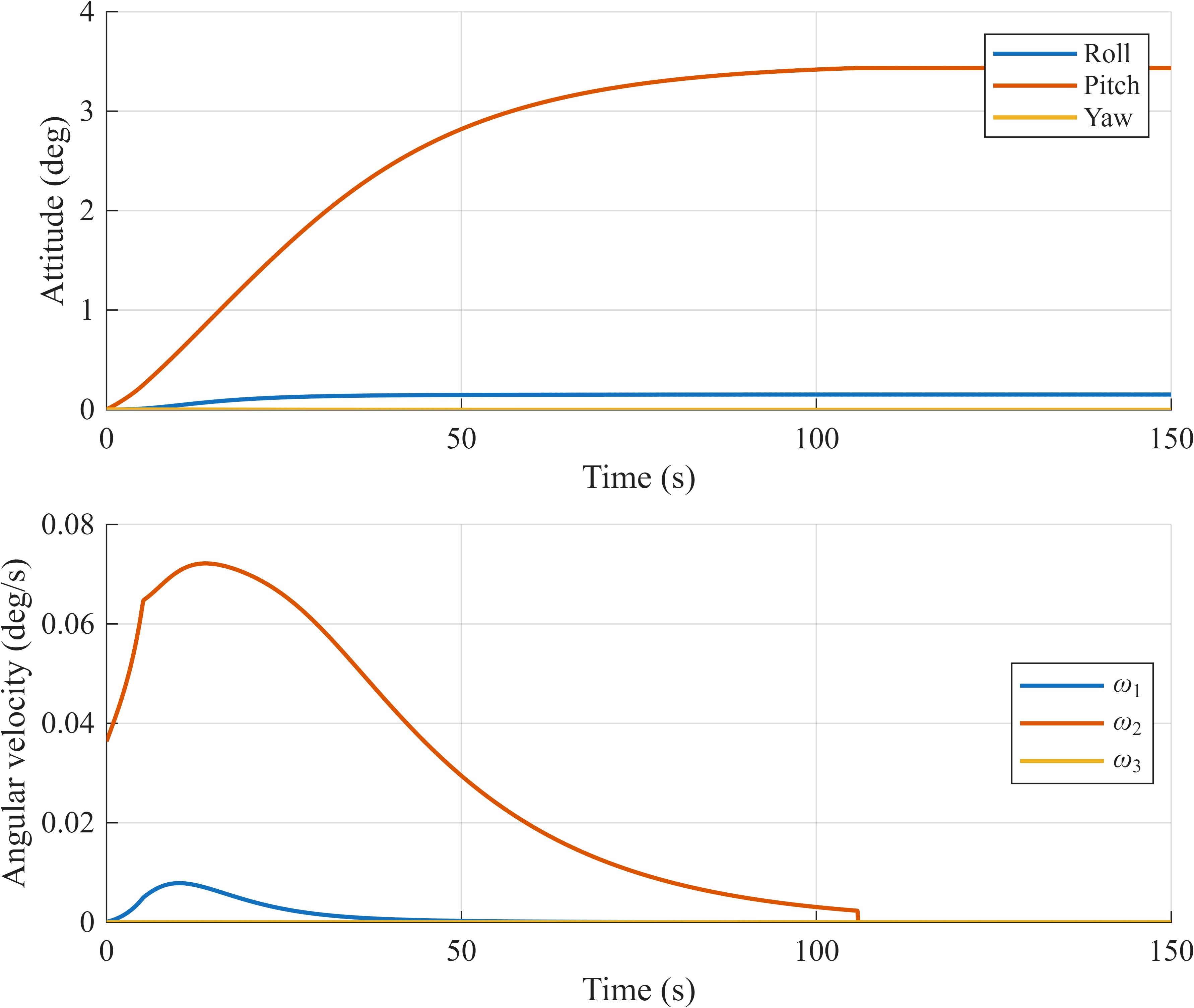}
        \label{fig:case1_base}
    \end{subfigure}
    \caption{Time History of Joint States, Base Attitude, and Base Angular Velocity (Case 1)}
    \label{fig:case1}
\end{figure}

Figure~\ref{fig:case1} depicts the system response when the decoupling constraint is applied to the yaw axis in Case 1. The base attitude plot clearly shows that the yaw attitude remains strictly at zero throughout the maneuver, while the roll and pitch attitudes vary, confirming that these unconstrained axes absorb the reactive momentum caused by the manipulator motion. The angular velocity of the base supports this evaluation: $\omega_1$ and $\omega_2$ fluctuate dynamically, whereas $\omega_3$ remains regulated at zero. 

\begin{figure}[!t]
    \centering
    \includegraphics[width=0.49\linewidth]{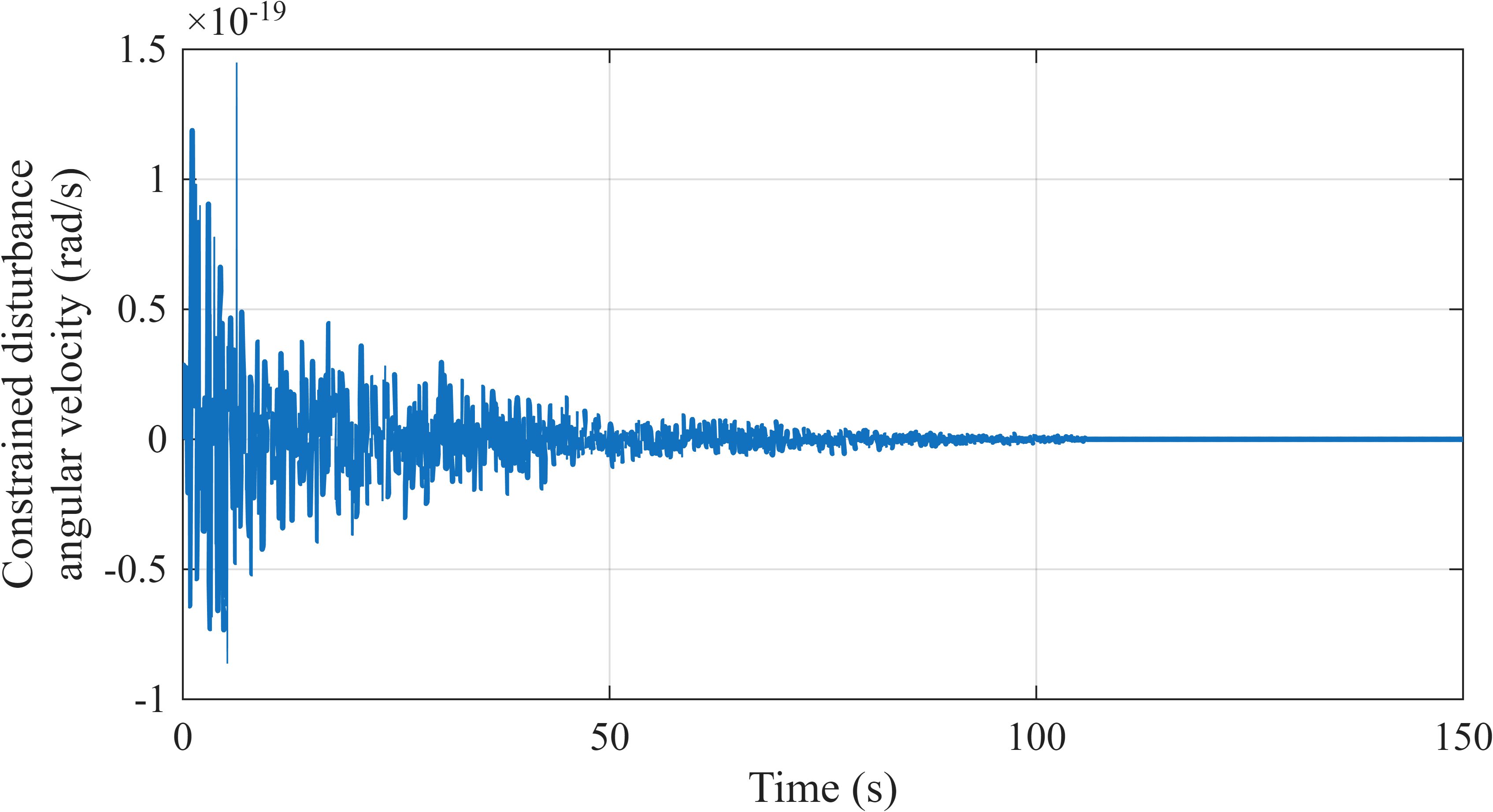}
    \caption{{Time History of} Constrained Disturbance Angular Velocity (Case 1)}
    \label{fig:case1_w_sub}
\end{figure}

The constrained disturbance angular velocity shown in Figure~\ref{fig:case1_w_sub} displays a numerical error bounded within a magnitude of $1.5\times 10^{-19}$~rad/s, validating exact mathematical decoupling via the null-space projection. Furthermore, the joint angles evolve smoothly from their initial states without sudden variations. The joint rate profiles reveal that Joint 5 initially exhibits a higher rate to drive the EE toward the target, with all joint rates successfully converging to zero at steady-state. By selectively decoupling only the designated critical axis, the manipulator utilizes the remaining DOFs to fulfill the operational task without encountering unnecessary workspace limitations.

\begin{figure}[!t]
    \centering
    \begin{subfigure}[b]{0.49\textwidth}
        \centering
        \includegraphics[width=\textwidth]{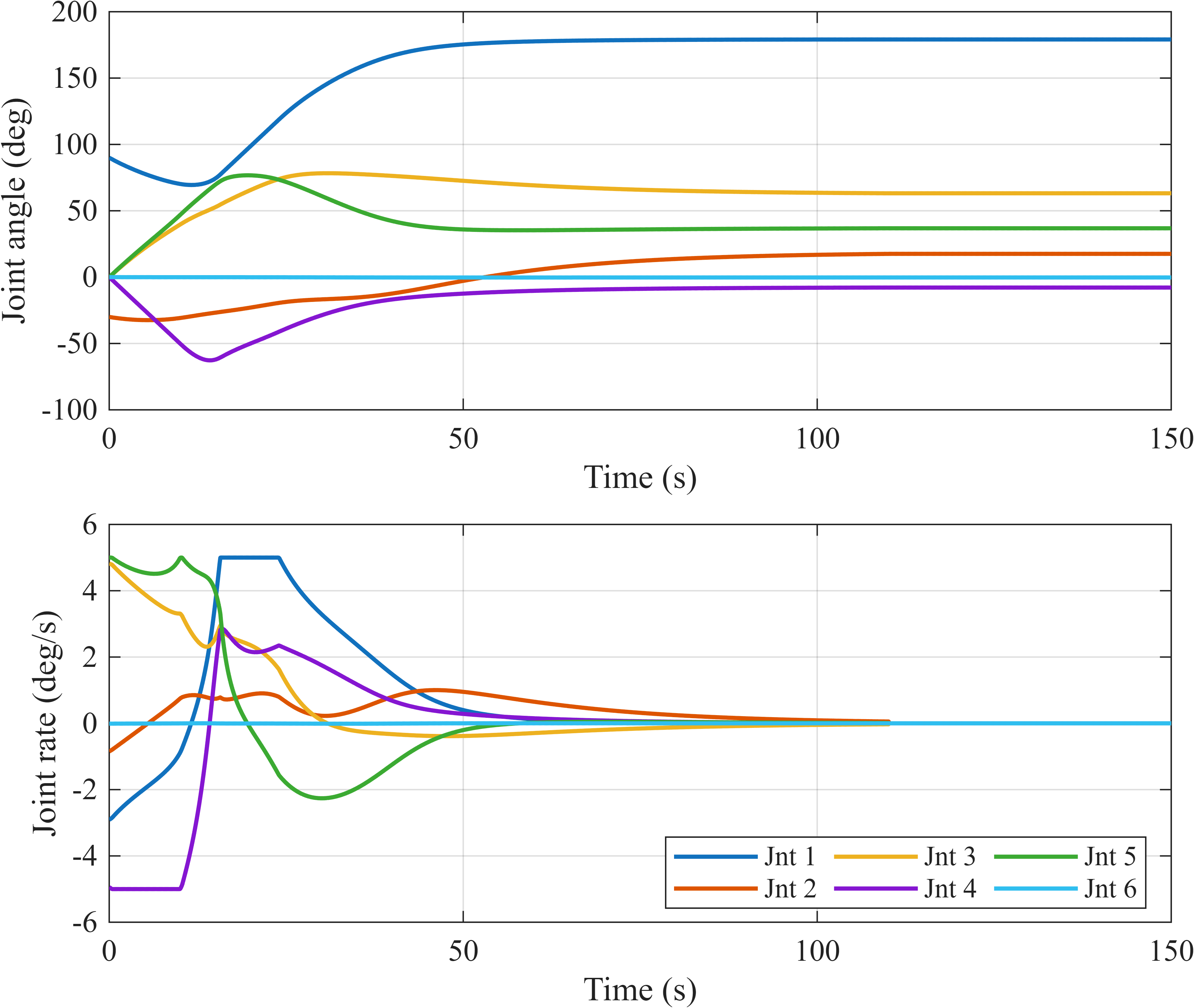}
        \label{fig:case2_jnt}
    \end{subfigure}
    \begin{subfigure}[b]{0.49\textwidth}
        \centering
        \includegraphics[width=\textwidth]{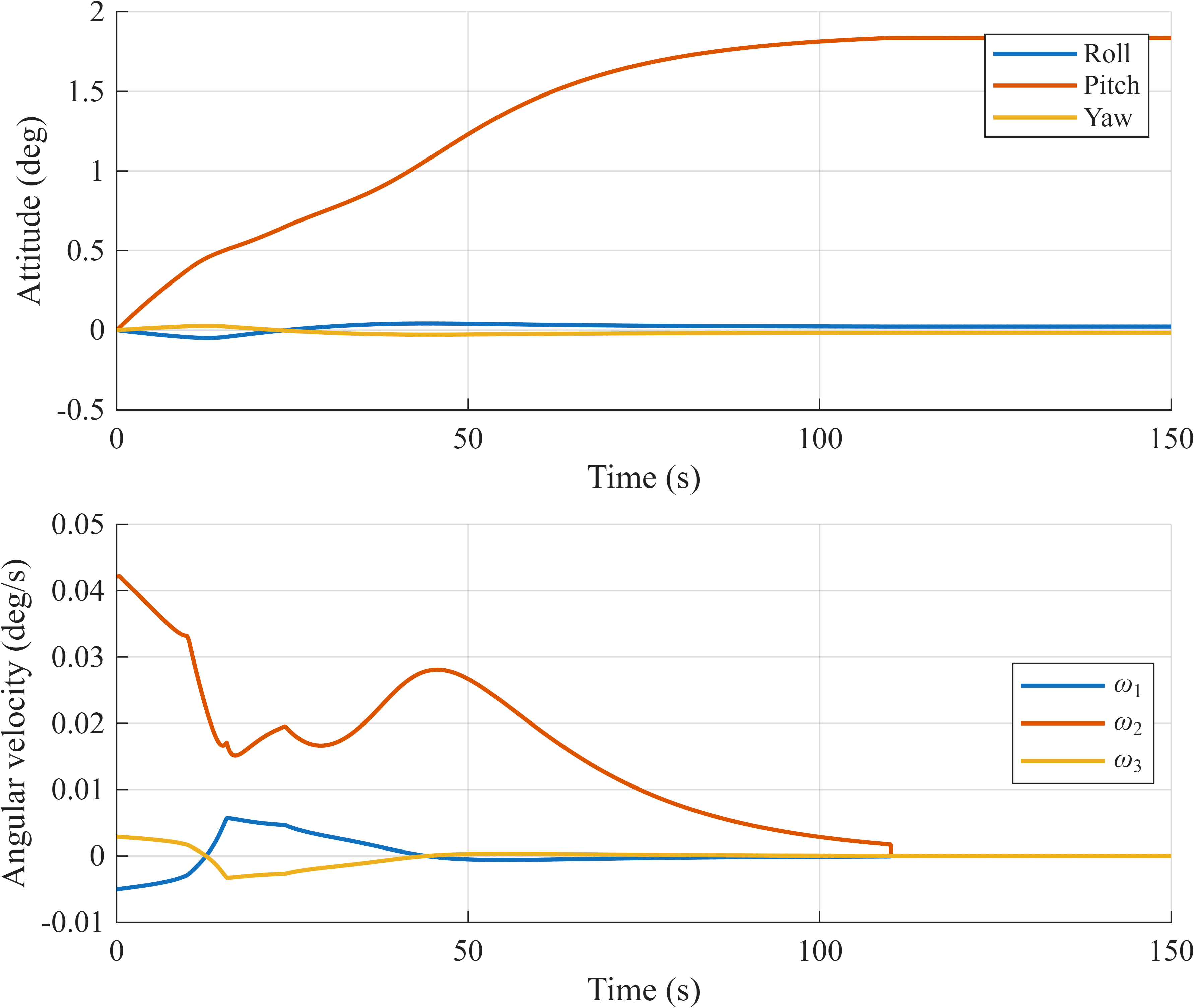}
        \label{fig:case2_base}
    \end{subfigure}
    \caption{Time History of Joint States, Base Attitude, and Base Angular Velocity (Case 2)}
    \label{fig:case2}
\end{figure}

\begin{figure}[!t]
    \centering
    \includegraphics[width=0.49\linewidth]{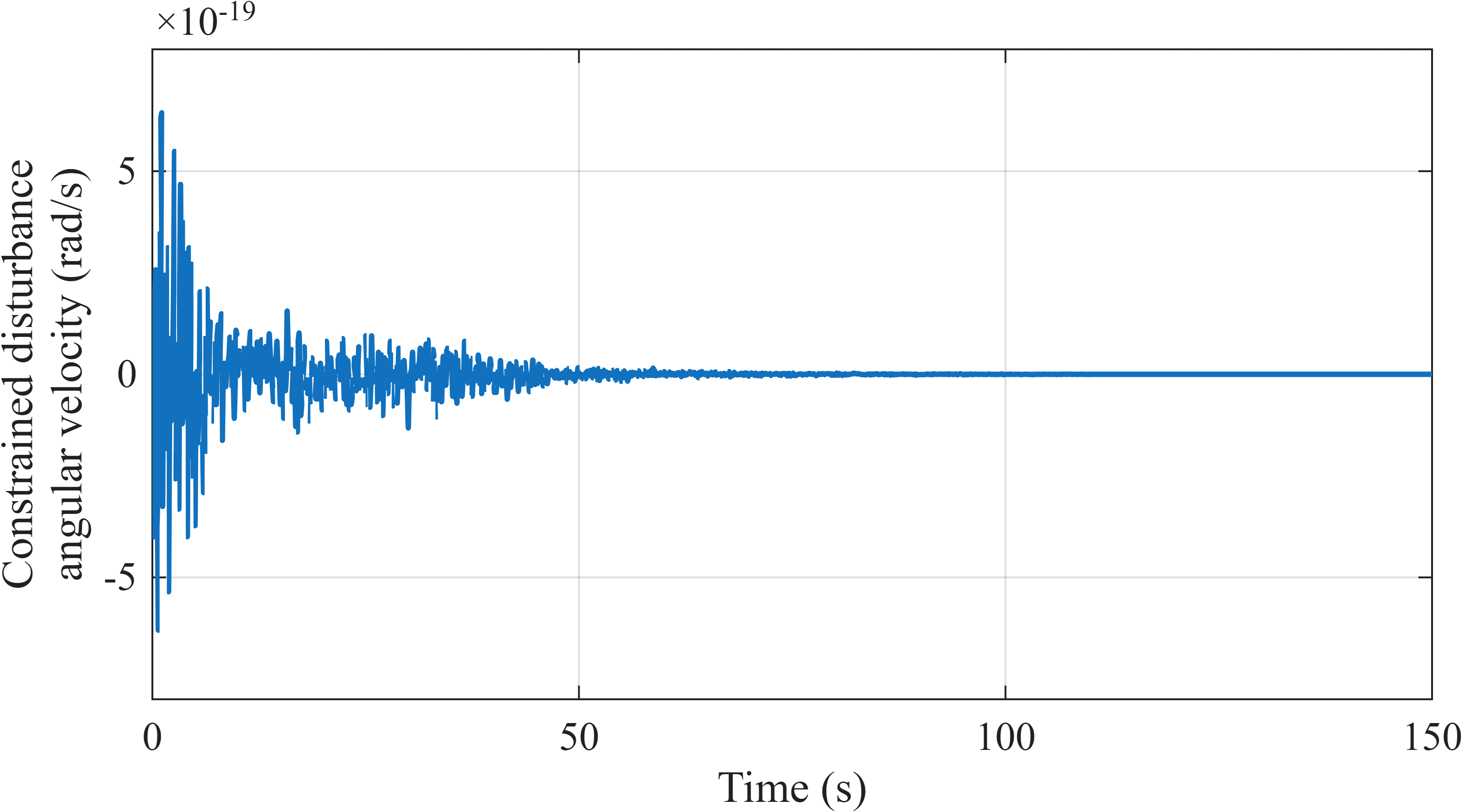}
    \caption{{Time History of} Constrained Disturbance Angular Velocity (Case 2)}
    \label{fig:case2_w_sub}
\end{figure}

The robustness of the generalized constraint matrix formulation is verified in Figures~\ref{fig:case2} and \ref{fig:case2_w_sub}. Unlike the principal-axis scenario of Case 1, the attitude and angular velocities across all three axes in Figure~\ref{fig:case2} are nonzero during the maneuver because the decoupling vector is intentionally misaligned with the base's principal body axes. However, the directionally constrained angular velocity is perfectly nullified, bounded within a magnitude of $7\times 10^{-19}$~rad/s, as shown in Figure~\ref{fig:case2_w_sub}. 

The complexity of protecting an arbitrary spatial direction is reflected directly in the joint maneuvers. Although Joints 1, 4, and 5 actively modulate their trajectories to compensate for the base momentum transferred along the $U_c$ direction, the joint rates are efficiently distributed by the optimization framework and safely settle to zero. This result highlights the flexibility of the proposed selective decoupling method, showing it can isolate disturbances along arbitrary spatial trajectories to easily accommodate highly asymmetric SMS configurations.

\begin{figure}[!t]
    \centering
    \begin{subfigure}[b]{0.49\textwidth}
        \centering
        \includegraphics[width=\textwidth]{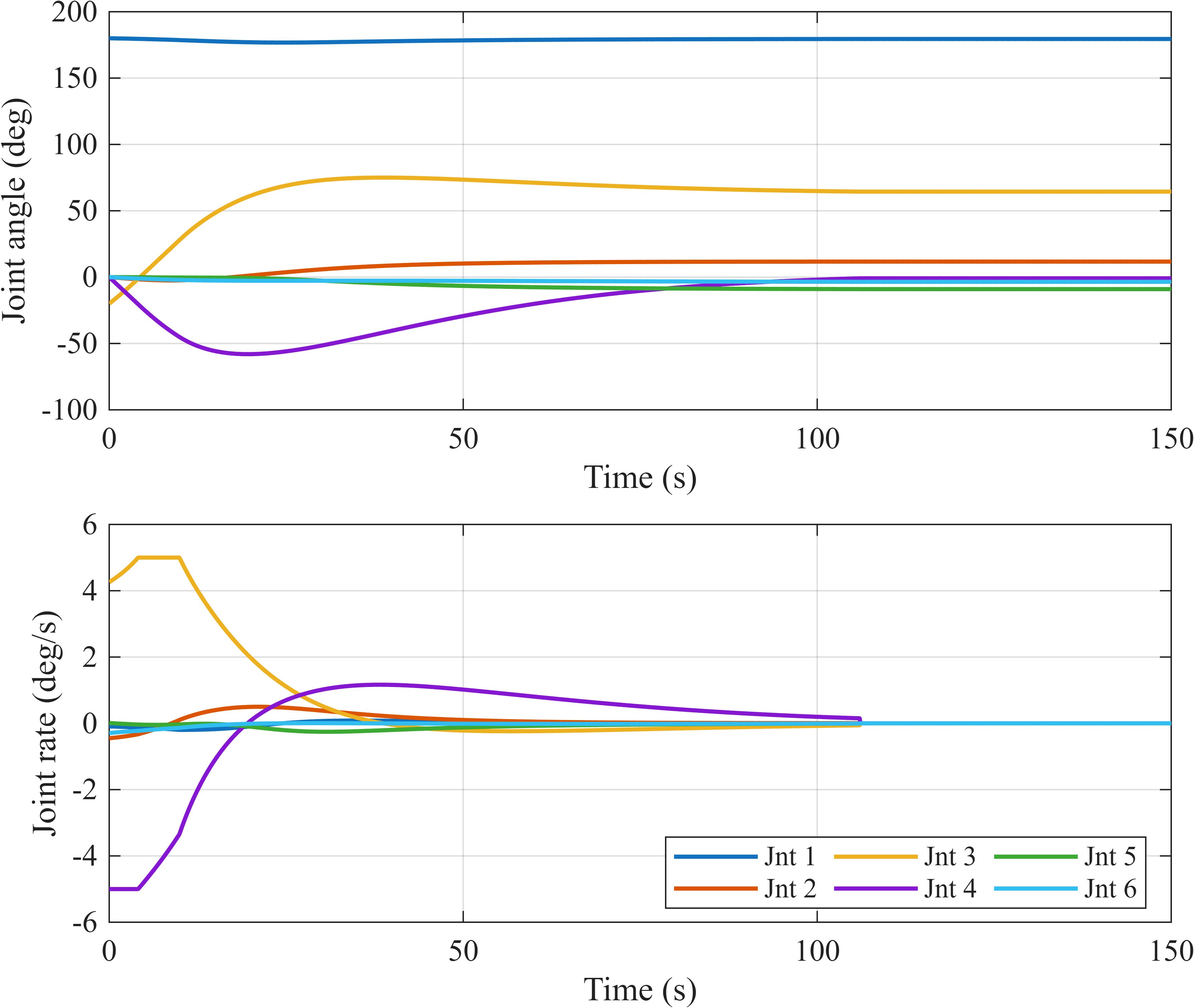}
        \label{fig:case3_jnt}
    \end{subfigure}
    \begin{subfigure}[b]{0.49\textwidth}
        \centering
        \includegraphics[width=\textwidth]{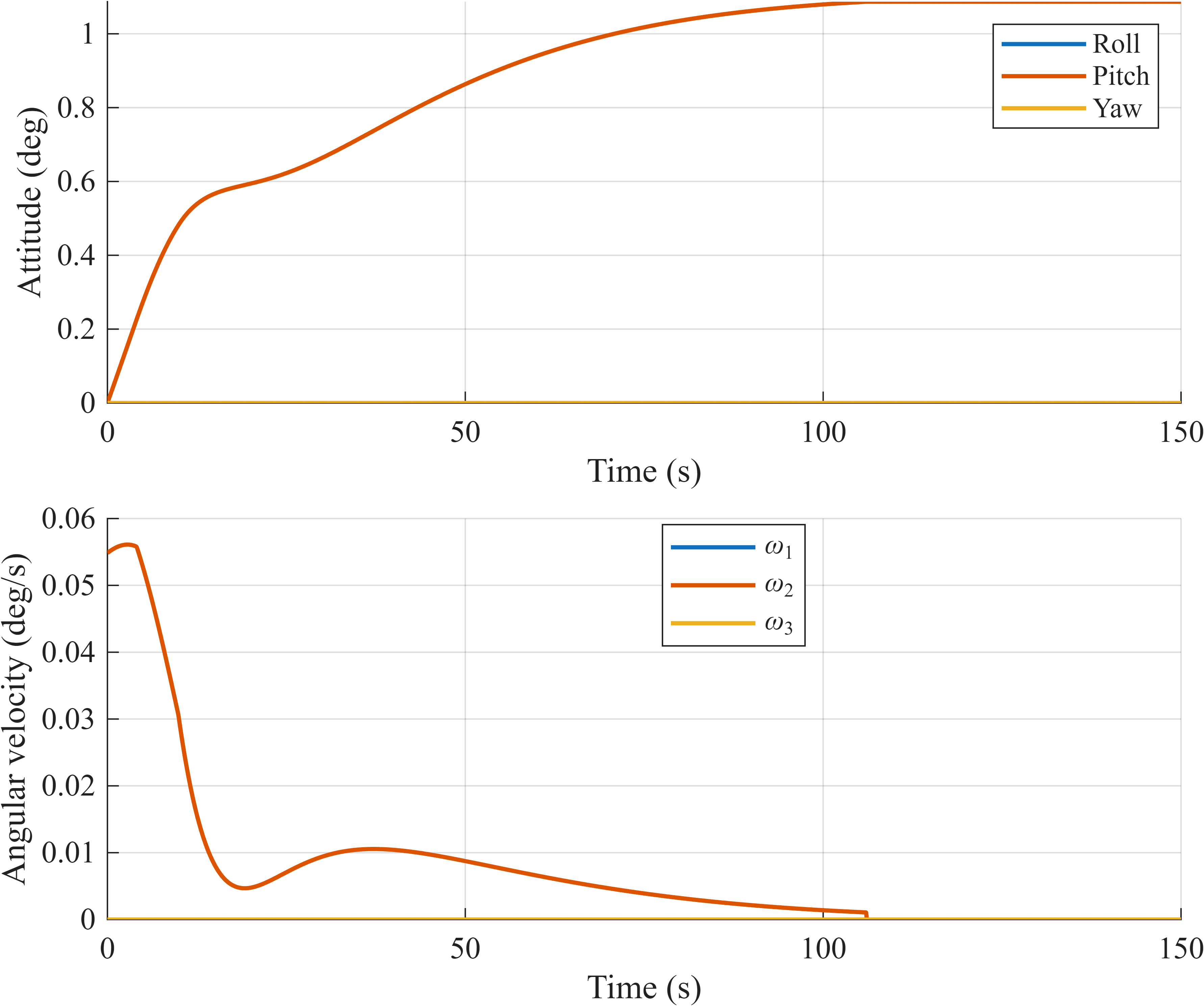}
        \label{fig:case3_base}
    \end{subfigure}
    \caption{Time History of Joint States, Base Attitude, and Base Angular Velocity (Case 3)}
    \label{fig:case3}
\end{figure}

\begin{figure}[!t]
    \centering
    \includegraphics[width=0.49\linewidth]{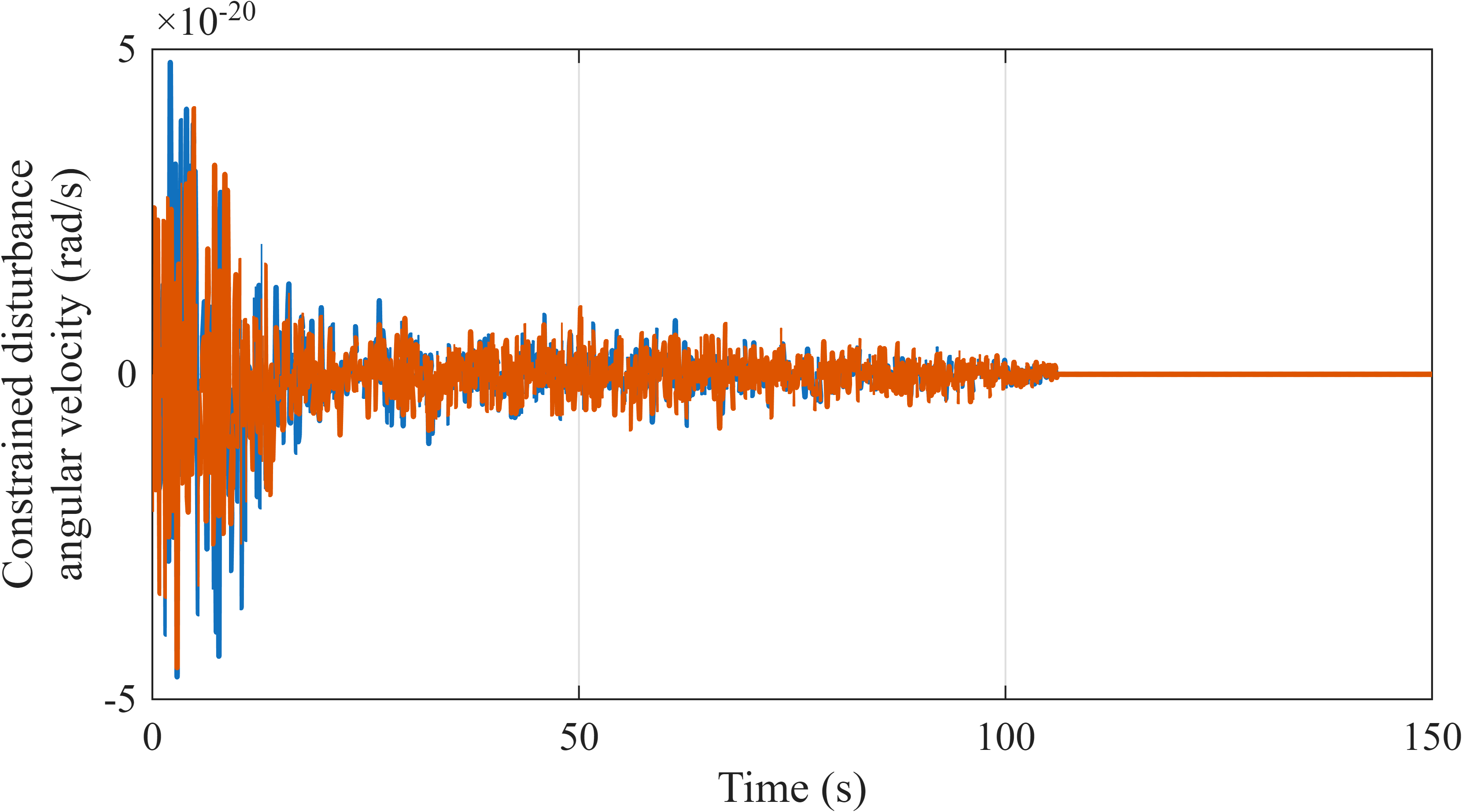}
    \caption{{Time History of} Constrained Disturbance Angular Velocity (Case 3)}
    \label{fig:case3_w_sub}
\end{figure}

Figures~\ref{fig:case3} and \ref{fig:case3_w_sub} illustrate the performance under a multi-axis constraint ($k=2$) designed to maintain antenna LOS tracking. The proposed framework successfully nullifies base disturbances along both the roll and yaw axes simultaneously. Consequently, the base attitude drift is isolated entirely to the unconstrained pitch direction. The resulting joint angle and joint rate profiles are smooth, demonstrating that the priority-based motion planning allocation successfully resolves joint rate commands that satisfy the primary EE tracking demands while satisfying the 2-DOF decoupling constraints.

In summary, the simulation results confirm that the proposed multi-axis selective decoupling framework successfully isolates the transferred momentum along specified mission-critical directions. This approach strategically suppresses undesirable base momentum along critical directions and redistributes it into non-critical axes, allowing the manipulator to maximize the feasible operational workspace.

\section{Conclusion}
This work introduces a multi-axis selective decoupling framework for free-floating space manipulator systems to address the limitations of conventional full-axis reactionless manipulation, which often leads to severe operational workspace reduction and kinematic singularities. The developed framework selectively isolates the base reaction along user-defined, mission-critical directions. By projecting the base motion into a directional constraint space, this methodology mathematically nullifies disturbances along the constrained directions, enabling the manipulator to leverage its remaining degrees of freedom for task-space missions. The effectiveness of this approach is validated through numerical simulation studies under three representative scenarios. The results demonstrate that the framework successfully nullifies momentum transfer to the base along a single principal body axis (Case 1), an arbitrarily defined spatial direction accommodating asymmetric spacecraft geometries (Case 2), and multiple axes to maintain specific pointing requirements (Case 3). In all cases, the framework mathematically decouples mission-sensitive directions, perfectly zeroing the constrained disturbance within a negligible numerical tolerance, while strategically redistributing the reactive momentum into non-critical axes to successfully execute the primary task-space mission. This strategy provides a flexible and adaptable solution for complex in-space servicing operations. Future work will extend this selective decoupling approach to dual-arm space manipulators, exploring collaborative momentum management and active disturbance compensation.

\section{Acknowledgment}
This material is based upon work supported by the Air Force Office of Scientific Research under award number FA9550-24-1-0600.

\bibliographystyle{AAS_publication}     % Number the references.
\bibliography{references}               % Use references.bib to resolve the labels.

\end{document}